\documentclass[preprint,12pt]{elsarticle}

\usepackage{graphicx}
\usepackage{amssymb}
\usepackage{amsmath}
\usepackage{soul}
\usepackage{xcolor}

\journal{ Biomedical Signal Processing and Control}

\begin{document}

\begin{frontmatter}


\title{{Proposing method to} Increase the detection accuracy of stomach cancer based on color and lint features of tongue {using CNN and SVM}}


\author[label1]{Elham Gholami, Instructor}
\author[label2]{ Seyed Reza Kamel Tabbakh\corref{cor1}, Assistant professor \corref{cor1}}
\author [label3]{Maryam kheirabadi, Assistant professor }
\address[label1]{Department of computer engineering, Neyshabur branch, Islamic Azad University, Neyshabur, Iran, gholami.elh@gmail.com}
\address[label2]{Department of computer engineering, Mashhad branch, Islamic Azad University, Mashhad, Iran, drkamel@mshiau.ac.ir}
\address[label3]{Department of computer engineering, Neyshabur branch, Islamic Azad University, Neyshabur, Iran, m.kheirabadi@iau-neyshabur.ac.ir	}
\cortext[cor1]{corresponding author: Seyed Reza Kamel Tabbakh}

\begin{abstract}
Today, gastric cancer is one of the diseases which affected many people's life. Early detection and accuracy are the main and crucial challenges in finding this kind of cancer. In this paper, a method to increase the accuracy of the diagnosis of detection cancer using lint and color features of tongue based on deep convolutional neural networks and support vector machine is proposed. In the proposed method, the region of tongue is first separated from the face image by {deep RCNN} \color{black} Recursive Convolutional Neural Network (R-CNN)  \color{black}. After the necessary preprocessing, the images to the convolutional neural network are provided and the training and test operations are triggered. The results show that the proposed method is correctly able to identifying the area of the tongue as well as the patient's person from the non-patient. Based on experiments, the DenseNet network has the highest accuracy compared to other deep architectures. The experimental results show that the accuracy of this network for gastric cancer detection reaches 91\% which shows the superiority of method in comparison to the state-of-the-art methods.
\end{abstract}


\end{frontmatter}


\section{Introduction}
\label{Sec:Introduction}

Cancer has been known as one of the main problems in human societies so that it has the most number of deaths after heart diseases~\cite{fuchs1995gastric}. Gastric cancer is the second common cancer in Asia~\cite{brunicardi2014schwartz}. About 800000 new cases of Gastric cancer are detected, and about 650000 people are died due to this disease, yearly~\cite{doherty2006current}.  Gastric cancer is the most common cancer in Iran that has been increasing during the last two decades rather than the other countries. Most of the infections of this disease are in the north and north-west of Iran. 

The most important risk factor of this disease is Helicobacter pylori bacterial infection that leads to ulcers and gastritis~\cite{li2014effects}. Moreover, several environmental and genetic factors affect gastric cancer.

The best method of gastric cancer diagnose is endoscopy, in which several biopsy removals of the wound margin are possible~\cite{[5]}. The detection precision of this method reaches 98\% by several biopsy removals. But it is an aggressive and costly method with anesthesia. If the method is not performed correctly, it has many side effects for the patient. Laparoscopy and gastroscopic examination are the other methods that help the disease diagnose by double-contrast imaging of the stomach include entering a camera into the throat and stomach~\cite{crew2006epidemiology}. Other examinations that are less common include barium swallow, X-ray imaging of the upper digestive system, and endoscopic sonography. But, if a person doesn’t have any symptoms of the disease, he/she doesn’t like to take such aggressive investigation methods. The new test of respiratory analysis is chipper, faster, and easier than the existing methods~\cite{xu2013nanomaterial}. The patients should blow on a pipe to perform such a test. Then the physicians check some sensors' reactions using the respiratory sample. Among the mentioned methods, the serologic methods, especially Urea Breathing Test (UBT), are more popular because of lower aggression and more reasonable cost. But these tests remain positive after 6 months to 2 years after treatment.
Incorrect diagnosis is the biggest obstacle to the treatment of gastric cancer patients in Iran~\cite{yarhusseini2014survival}. Unfortunately, there is no common test in Iran for screening gastric cancer patients. Even the disease in some patients with symptoms like stomachache, weight loss, dysphagia, and vomiting for a long time is not diagnosed. If gastric cancer is diagnosed early, it is treatable, but if the disease is progressed, the treatment probability decreases. Therefore, based on the high prevalence of gastric cancer, some patients refer in advance stages of the disease, the existence of appropriate diagnose tools for early diagnosis, determine its importance in the early stages for the amount of the patients' survival~\cite{delpisheh2014smoking}. Usually, treatment is performed by a combination of stomach surgery, chemotherapy, and radiotherapy with some auxiliary treatments~\cite{yonemura2010surgical}. The most usable treatment for gastric cancer is gastrectomy surgery (tumor removal), especially the cancer is in the early stage.

Chinese traditional medicine is a natural and comprehensive health care system with 3-5 thousand years of antiquity~\cite{sarebanha2016comparison}. Iranian traditional medicine is a fine and valuable heritage that its antiquity returns to 5 thousand years before Christ in the Achaemenid era~\cite{smith1980avicenna}. The emphasis of Iranian traditional medicine that is a combination of Persian culture, community, and religious sciences, is on the balance in the human body. It could be popular in today's technology age because of its effectiveness, help to human health, and human suffering reduction. Chinese traditional medicine has a long history of different diseases treatment in eastern Asia countries. Moreover, it is known as a complementary system and medical alternative in western countries.

Today, traditional medicine is not only popular among people but also many of the scientific and medical centers around the world from east to west have researched and trained it. Moreover, the World Health Organization (WHO) encourages health policymakers of different countries to use the medicine and combine it with the common medicine. The scientific departments of this medicine are increasing every day around the world. The WHO strategy from 2014 to 2023 has explained that Chinese traditional medicine will be expanded in the universities of more than 100 countries, and it will be converted into an industry~\cite{world2013traditional}. 

Despite the widespread use of Chinese traditional medicine in China and western countries, the precise scientific observations of its effects are limited. The disease diagnosis study through Chinese traditional medicine is difficult for the researchers due to the difference of its ideas with modern western medicine~\cite{chan2010interactions}. In Chinese traditional medicine, the disease diagnose is based on the obtained information from four diagnosis processes of seeing, hearing, smelling, and touching. Mostly, the diagnoses are based on taking the pulse and examining the tongue.

Investigation of the tongue surface is one of the most important diagnostic methods of the diseases in Chinese traditional medicine that is used extensively. The tongue is so important in Chinese traditional medicine because it is in palate far from external and environmental factors~\cite{chiu2000novel}. They divide the tongue into four parts of the tongue tip, tongue margin, center of the tongue, and end or tongue growth. Figure 1 shows the relationship between different parts of the tongue and internal organs~\cite{wang2011basic}. Tongue tip shows the pathologic changes of the heart and lungs. Two sides of the tongue show liver and bladder sac changes. The tongue center is related to pathologic changes of stomach and spleen, and the end of the tongue is related to changes of kidneys, intestines, and bladder.

\begin{figure}[h]
\centering\includegraphics[width=0.5\linewidth]{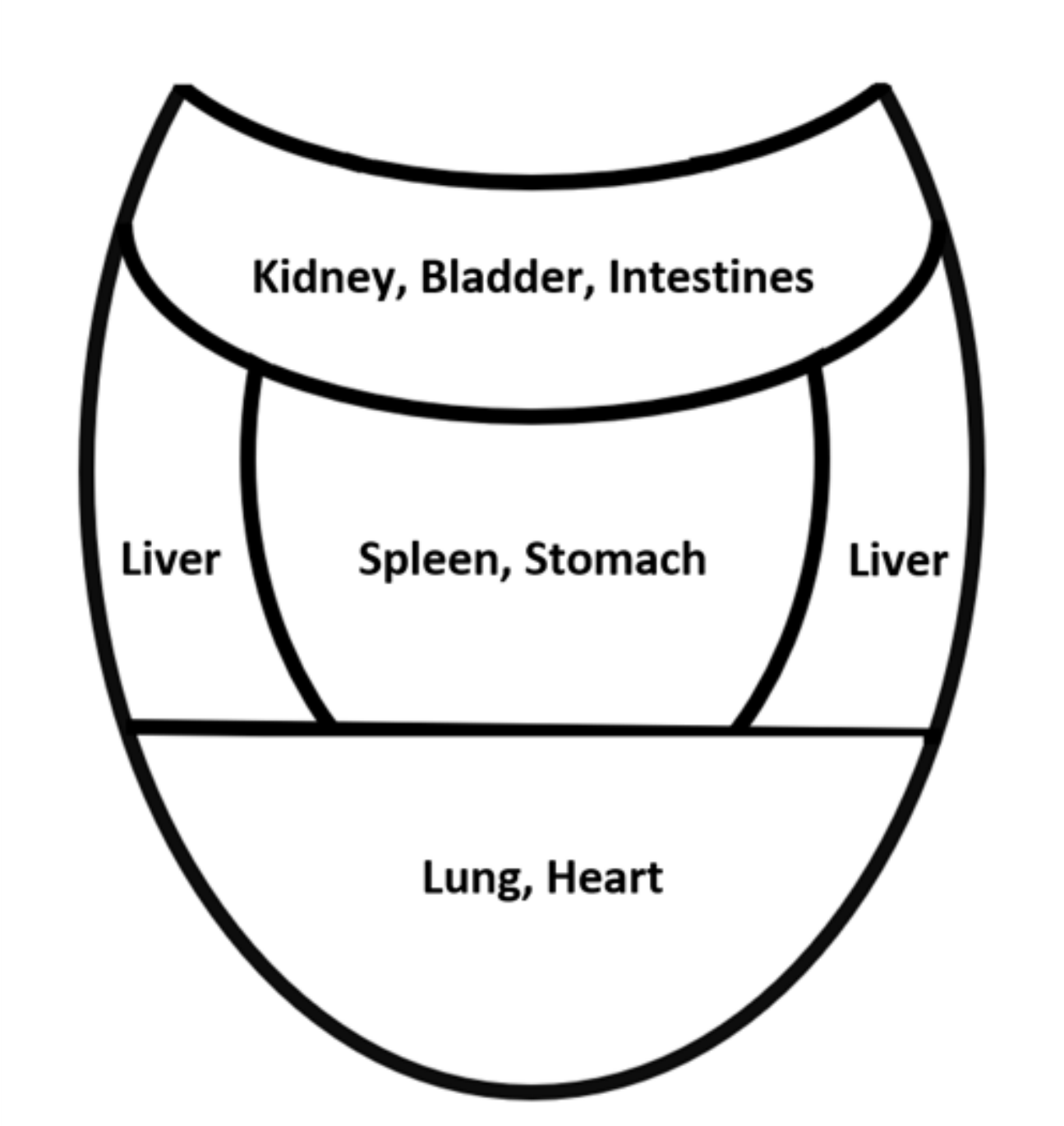}
\caption{The relationship between different parts of tongue and internal organs}
\end{figure}

The physicians diagnose different diseases according to the shape and the size of the tongue, cover on the tongue, the tongue color, and moisture on the tongue~\cite{maciocia1995tongue}. A healthy language is oval and pink with a very thin cover. Some of the disease symptoms in tongue include color change, thickness increment, its cover change, change of its geometric shape, and creation of grooves on the tongue.

{Now,} General gastric cancer and helicobacter pylori infection diagnosis methods {are} \color{black} can be \color{black} divided into aggressive and non-aggressive~\cite{peter1999evaluation}.

\subsection{Aggressive Methods}

\begin{enumerate}
    \item Endoscopy: a method with anesthesia and high cost that sometime when active or recent bleeding of the patient, or when he/she is under anti-secretion treatment, its answer is false negative.
    
    \item Polymerase Chain Reaction: in this method, helicobacter pylori are diagnosed in small tissue samples, including some bacteria. The fault of this method is the DNA of bacteria that can be in stomach mucus from dead bacteria after treatment and results in false-positive tests~\cite{blaser2006we}.
    
    \item 	Rapid Urease Test: this method is one of the bases of infection diagnosis utilizing helicobacter pylori capability in the generation of a high amount of urea.  This method is faster and cheaper than other methods like tissue culture and investigation. But its sensitivity depends on organism density, meaning that when the number of organisms is low, the test sensitivity decreases up to 30\%.

\end{enumerate}

\subsection{Non-aggressive methods}

\begin{enumerate}
    \item Serological tests: it is a cheap method, but since the H.pylori strains in various geographic areas are different, not using the local anti genes of each area in laboratory diagnostic kits results in sensitivity reduction of the tests. Moreover, many false-positive tests have been observed due to the generated antibodies of other infections and cross-reaction with this section~\cite{dore2004novel}.
\item Urea Breathing Test: this method is based on Helicobacter pylori urease enzyme activity that is a time-consuming process and requires expensive equipment. Moreover, existing of other positive urease bacteria in the mouth and stomach holes leads to false-positive results.
\item Determination test of Helicobacter pylori antigen in feces: this method is utilized for the detection of helicobacter pylori antigen existence in feces samples using the ELISA method that is more expensive than other tests~\cite{pourakbari2011evaluation}.

   \end{enumerate}

Cover on the tongue shows physiologic and Pathologic changes in digestive organs, especially the stomach~\cite{liu2015metabonomic}. The reports of clinical researches show that tongue appearance provides essential information for physicians to diagnose, treat, and prediction of chronic diseases of stomach, gastrointestinal ulcers, gastric, and colon cancers~\cite{jiang2012integrating}. Since the researchers try to find easy and cheap methods for early diagnose of cancer, they can use the relation between coating metabolic symptoms and cover of tongue surface to diagnose chronic diseases like gastric cancer.

\subsection {Roadmap}
The structure of this paper is as follows. Section 2 includes the introduction and investigation of the previous related works. The proposed method has been explained in section 3. The experimental results have been presented in section 4. Finally, section 5 concludes the paper and proposes some ideas as future works.

\section{ Literature Review}
Today, different methods have been proposed to diagnose cancer and various kinds of diseases such as liver disorders, heart diseases, hepatitis disorders et al. utilizing artificial intelligence and different algorithms in the fields of machine learning~\cite{neshat2013survey, neshat2014diagnosing, neshat2009feshdd, neshat2008designing}, evolutionary methods, and neural networks~\cite{neshat2010hopfield, mandal2014comparative,neshat2009designing,neshat2012hepatitis}. Generally, each of the proposed methods has some advantages and disadvantages. In this section, the advantages and disadvantages of the recent researches have been investigated.

Zhang et. al, (2017) in~\cite{zhang2017diagnostic} have proposed Diabet diagnoses based on tongue standard images using the SVM algorithm. They have taken the tongue images of 296 Diabetic patients, and 531 healthy cases using tongue digitalize imaging tools. The shape and coating of the tongue in the images are different, and each of them is divided into different color spectrums. To this aim, Division-Merging and Chrominance-Threshold algorithms have been used. The SVM algorithm has been used for classification of the tongue color and texture, that its Kernel parameter has been optimized using the Genetic Algorithm (GA). Hu et. Al (2015) in~\cite{hu2015variations} have studied the relation between the tongue cover observations and mouth bacteria in gastric cancer patients. Their statistical society includes 74 cancer patients and 72 healthy people. They use tongue images of the fasting people using the DSol-B tool that is an imaging tool with the capability of thickness and coating analysis of the tongue. For sample gathering, first, the patients wash their mouth with buffered saline. Then samples are taken from their tongue by scrubbing. They show that tongue cover with high thickness represents less diversity of the bacteria rather than thin tongue in gastric cancer patients.

Cheng \color {black} et al.\color{black}, (2015) in~\cite{hu2016color} have designed a framework to diagnose tongue images automatically. They provide tongues images using a smartphone with a D50 imaging standard. The images have been classified using the SVM algorithm for prediction of light conditions and color correction matrix based on two states of with and without flash. In the framework, cracks and coating of the tongue have been detected in the last stage. Han et. al, (2014) in~\cite{han2014tongue} have been researched on the relation of tongue cover microbiome and colon cancer. They use the DSol-B imaging system to take the tongue images of 47 patients and 45 healthy people. Then by determining next-generation sequence, the tongue thickness is detected, and the results have been analyzed by SPSS. The results show that the tongue thickness of the patients of colon cancer is more than healthy people.

According to the diagnosis of gastric cancer form the tongue images that is a two-class problem (\color{black} class 1 \color{black}: the disease existence detection, and class 2: the disease non-existence detection or healthy state). The average of true diagnosis precision of the two classes is 50\% random. Moreover, according to the importance of the disease subject, the expected average precision  \color{black} is \color{black} more than 90\%. Therefore, having a proper method and algorithm to reach this precision is essential. To this aim,  \color{black} we have reviewed the \color{black}  hard and complex problems of diagnosing and classification of the images with more than  \color{black}50 \color{black} classes, to have a deeper and better view of the challenges, the proposed algorithms, and the problems' solutions to diagnose the disease in two classes.

\section {The Proposed Method}
In this section, the proposed method has been explained to increase the precision of gastric cancer diagnosis using the features of coating and color of tongue based on convolutional deep neural networks and support vector machine. Figure 2 illustrates the diagram of the proposed method.

\begin{figure}[h]
\centering\includegraphics[width=\linewidth]{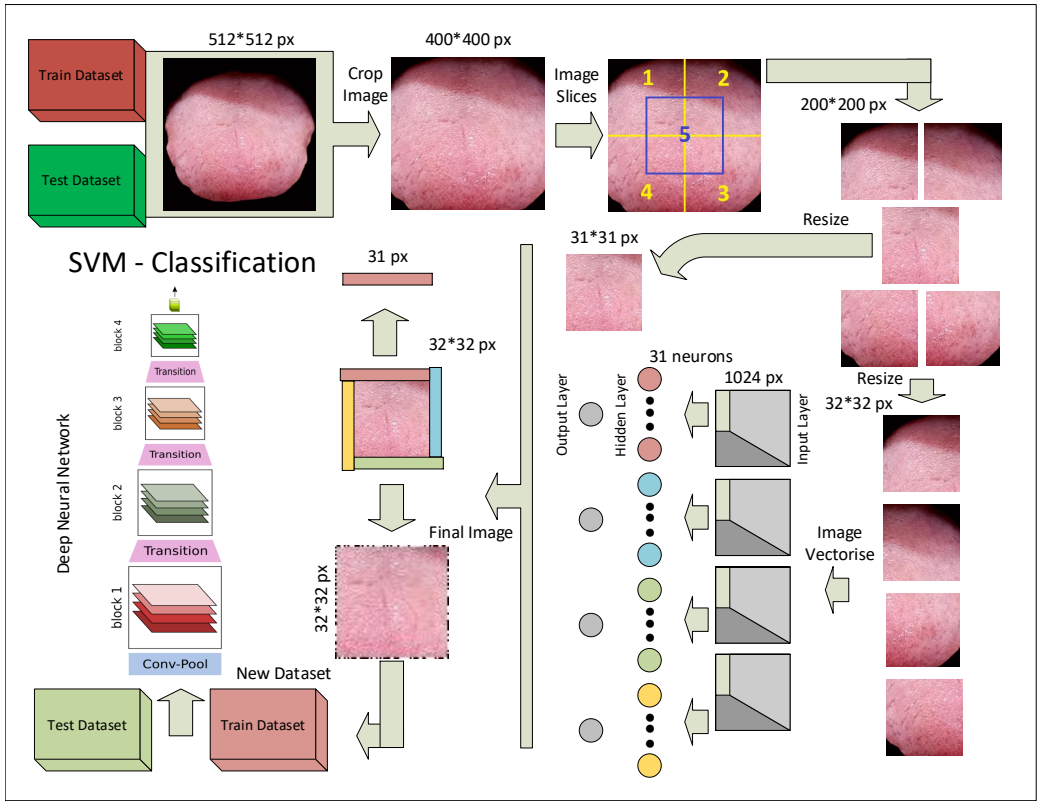}
\caption{The relationship between different parts of tongue and internal organs}
\end{figure}

\subsection{Preprocessing}
Dataset of tongue images have been prepared in two classes to diagnose gastric cancer by human tongue image. The primary raw image dimensions are 2988×5312 pixels. A sample of an example of the raw image (primary) of the healthy images' class has been depicted in figure 3.

\begin{figure}[h]
\centering\includegraphics[width=0.5\linewidth]{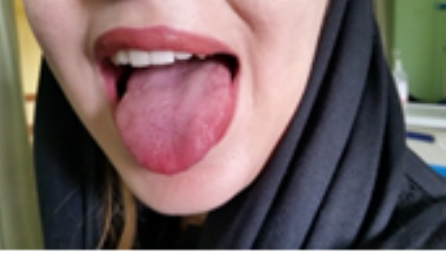}
\caption{Primary image sample of the normal images class}
\end{figure}

As the figure shows, the primary image is the tongue image with head and face image. RCNN deep neural networks have been utilized to separate the tongue area from the primary image. The neural networks are so powerful for things detection and specification of their area points in the images. So, RCNN deep neural network is one of the best methods to gain high precision for the detection of tongue area when there is a dataset with sufficient information of tongue area.

To this aim, pre-training has been used, and the trained neural network of AlexNet has been selected. At first, the images of the two classes are divided into training and test images. Then, the training images with tongue location are entered into the network input, and the network is trained. The test results are very good so that the error of tongue area detection is very low. Figure 4 shows an example of tongue area detection by RCNN deep neural network in the test stage. The image with a yellow tongue is the result of detection by RCNN, which shows high precision of tongue area detection.

\begin{figure}[h]
\centering\includegraphics[width=0.5\linewidth]{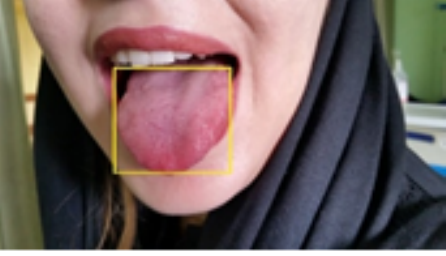}
\caption{Tongue area detection by RCNN deep neural network}
\end{figure}

After the detection of the tongue area from the primary images using RCNN, we have the points value and location of 4 vertices of the tongue area rectangle. So, we can consider an image with the dimensions like the dataset images as a reference image. Then we can consider a rectangular area of the reference image as the standard location of the tongue image. After consideration of the reference model image, we have the points value and location of 4 vertices of the tongue area rectangle. If the images for gastric cancer diagnosis are called the animated image, we can adapt the animations with the reference image and maximize their similarities. It results in the similarity of location and shape of the tongue in all the images of the dataset and locating them in the standard location. Moreover, it leads to the higher precision of training of the CNN deep neural network. Also, these three stages are used to diagnose gastric cancer from the test images: detection of the tongue location, matching the test image to the reference model, and diagnosis by the proposed method through the matched image.

Because one camera has been used to take the images and the animated and the reference images are homogeneous, the similarity criteria for matching the images is minimum mean square. Moreover, Affine conversion has been utilized because it includes a combination of base conversions of transfer, rotation, scale change, and convolution.

This conversion has six parameters. So, we should find the parameters for each image in the training images set. It is repeated for the test images, too. The equations of the considered adaptation have been presented. In these equations, $i$ shows the number of the image, and j is the number of the considered point in the i$^{th}$ image. Each image has 4 corresponding points with the rectangular area of the tongue. The point has been shown with $\overrightarrow{y}$. Each point has a component alongside $x$ length and another component alongside y width. Also, $r$ is the reference image index. Finally, there is a $T_i$ conversion for each image i that has six parameters of $t_{11},t_{12},t_{13},t_{21},t_{22},t_{23}$. Each corresponding image can be adapted and changed to the reference model. The conversions are performed based on Equation 1:


\begin{equation}
\begin{split}
T &=
\begin{bmatrix}
t_{11} & t_{12} & t_{13}\\
t_{21} & t_{22} & t_{23}\\
t_{31} & t_{32} & t_{33}
\end{bmatrix} ~~~~  \overrightarrow{y_{I,J}}= < x_{i,j}, y_{i,j}>
~~~~ \overrightarrow{y_{r,J}}= < x_{r,j}, y_{r,j}>\\
 C_{LMS}=& \sum_{j=1}^4 |\overrightarrow{y_{I,J}}-T_i \begin{bmatrix}
 \overrightarrow{y_{I,J}}\\
 1
 \end{bmatrix}|^2 \\
C_{LMS}= &\sum_{j=1}^{4} (x_{i,j}- t_{11}x_{r,j}-t_{12}y_{r,j}-t_{13} )^2 + (y_{i,j}- t_{21}x_{r,j}-t_{22}y_{r,j}-t_{23} )^2\\
\frac{\delta C_{LMS}}{\delta t_{11}}=&0\\
\sum^4 _{j=1}& -2 (x_{i,j}- t_{11}x_{r,j}-t_{12}y_{r,j}-t_{13})x_{r,j}=0        \\
t_{11}\sum_{j=1} ^4 & x_{r,j}^2 + t_{12}\sum_{j=1} ^4 y_{r,j} x_{r,j}+ t_{13}\sum_{j=1} ^4 x_{r,j} = \sum_{j=1} ^4 x_{i,j}x_{r,j} \\ 
\end{split}
\end{equation}

If the derivative of the function is taken in terms of each of the parameters and set to zero, we have a set of grade 2 equations like equation 2 that the conversion parameters are unknown. The unknown and the conversion parameters are obtained by solving the equations set.


\begin{equation}
\begin{bmatrix}
a & b & c & 0&0 &0\\
[~] & [~] & [~] & 0&0 &0\\
[~] & [~] & [~] & 0&0 &0\\
0 & 0 & 0 & [~] & [~] & [~]\\
0 & 0 & 0 &[~] & [~] & [~]\\
0 & 0 & 0 & [~] & [~] & [~]\\
\end{bmatrix}  
\begin{bmatrix}
t_{11} \\
t_{12} \\
t_{13} \\
t_{21} \\
t_{22} \\
t_{23} \\
\end{bmatrix}  
=
\begin{bmatrix}
d \\
[~] \\
[~] \\
[~] \\
[~] \\
[~] \\
\end{bmatrix}  
\end{equation}

Finally, all the points of each animated image in the training images set should be multiplied by the corresponding reverse conversion of that image. After interpolation, the new image that is adapted with the reference image is obtained. It is performed for the test images, too. The conversion is performed for a special point of j in image i, and the new position of that point is obtained $\overrightarrow{x}$, the final position of that point will be gained after interpolation. The conversion is performed for all of the points.

\begin{equation}
\overrightarrow{x_{i,j}}=T^{-1}_i \overrightarrow{y_{i,j}}
\end{equation}

The general diagram of the proposed method to detect the tongue area, adapt with the reference model, and cutting have been shown in figure 5.

\begin{figure}[h]
\centering\includegraphics[width=\linewidth]{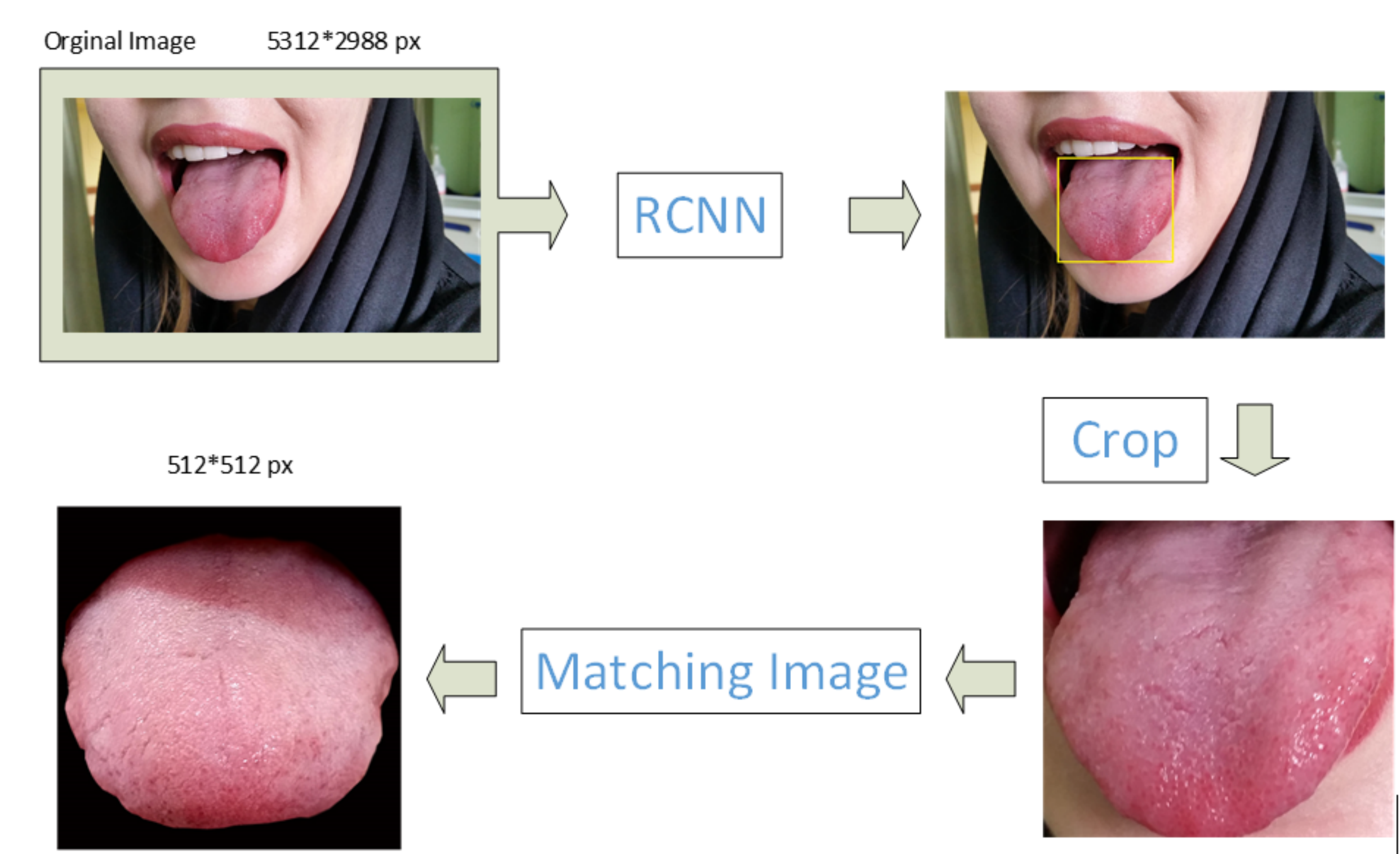}
\caption{Tongue area detection diagram and matching it with the reference model}
\end{figure}

The resulted images from the previous section are organized after preprocessing and detection and adaptation of the tongue area. Finally, the numbers of images of the healthy and patient classes are 700 and 870, respectively, and the total number of images is 1570. 200 images have been selected as the test images from each of the classes, randomly, and the total number of the test images are 400. Hence, the number of training images in the healthy and patient classes are 500 and 670, respectively, and the total number of the training images is 1170. Since providing a dataset of images is so difficult, the total number of images is not high. To have an acceptable number of the training data, the selected test images for each class is 200. All the images are in color with dimensions of 512×512 pixels. Figure 6 shows an example of a healthy image.

\begin{figure}[h]
\centering\includegraphics[width=0.3\linewidth]{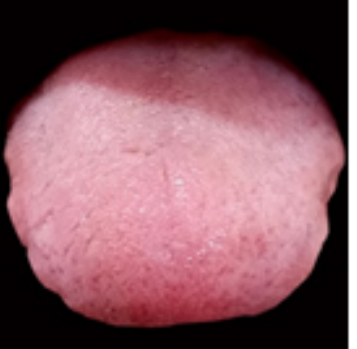}
\caption{An image sample of the normal images class}
\end{figure}

At first, this dataset of images has been trained and tested pre-training on different deep neural networks like AlexNet, ResNet, and Densnet. In the best state, the average classification precision of the test images in the two classes is 70\%. A method and an algorithm have been proposed to improve the classification precision that lead to enhancement of the average classification precision to 90\% for two classes. It will be explained in the results section.

\subsection{ The Main Processing}
The margins of each image are black in all the images. They have common information and a negative effect on classification. The margins are cropped by the written program automatically. Then the set of images is updated. Indeed, 56 pixels of each side of the image are cut that results in the new images with the dimensions of 400×400. Figures 7 and 8 show an example of the cut image.

\begin{figure}[h]
\centering\includegraphics[width=0.3\linewidth]{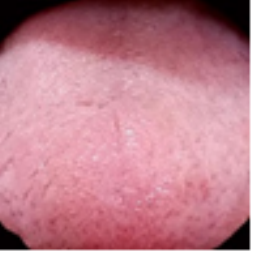}
\caption{Cropped image sample with new dimensions of 400×400 pixels}
\end{figure}

Then each of the new images is divided into 5 information areas of the tongue. Figure 6 shows an example of the image that the divided areas are determined on it.

\begin{figure}[h]
\centering\includegraphics[width=0.3\linewidth]{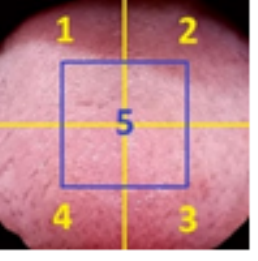}
\caption{Division of a tongue image into five regions}
\end{figure}

As the figure shows, area 5 is in the center of the figure that it has shared areas with the other four areas. So, it is predicted that area 5 has a higher information load. The experiments only using area 5 as the set of training and test images show the average classification precision improvement of 5\%. So, the assigned weight of area 5 is higher than in other areas that lead to the generation of the next stages of the proposed method. In figure 9, different areas that are cropped automatically with a written program are observable. The dimension of each area of the image is 200×200 pixels. After cutting, each area updates the images dataset, and different 5 datasets of images are created for each area.

\begin{figure}[h]
\centering\includegraphics[width=0.6\linewidth]{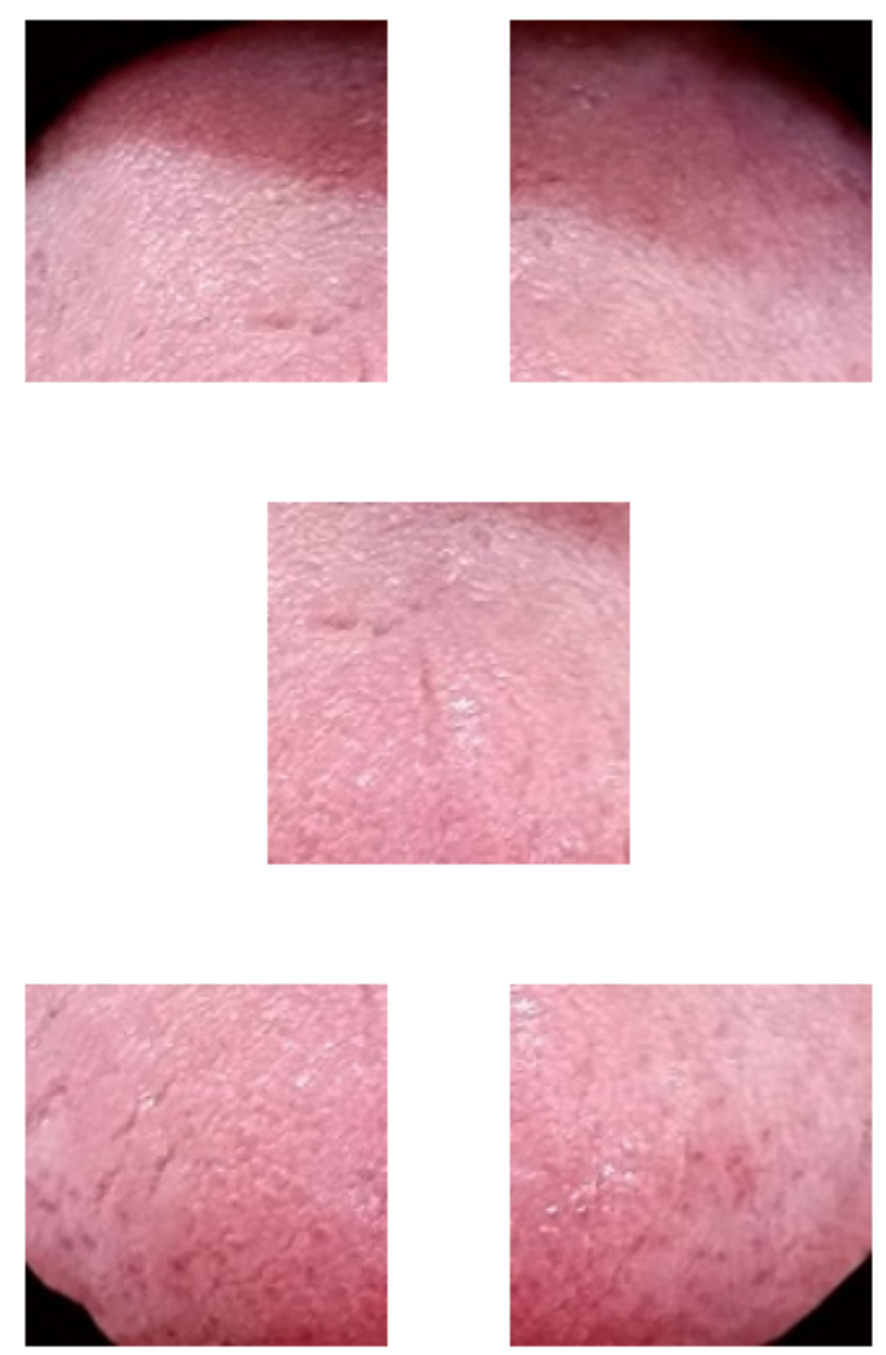}
\caption{An example of cropped images of the reference image in five different areas with dimensions of 200×200 pixels}
\end{figure}

The aim of the next steps of the proposed algorithm is the creation of a combined image of the 5 areas of the tongue image that the highest weight is for the fifth area. Moreover, good and effective features of areas 1 to 4 are extracted and combined with images of area 5. To this aim, at first, we reduce the dimensions of all the images of the dataset to 31×31. A separate algorithm has been applied to each of the images of areas 1 to 4, which is explained in the following. 

The algorithm of the image set of area 1 will be explained and repeated for areas 2 to 4. At first, the dimensions of area 1 images are reduced to 32×32 pixels, then channel one of the three color channels of each image is vectorized. So, each image dimension in the new dataset will be 1024 pixels. Now we configure a two-layer artificial neural network with 31 neurons in the hidden layer and 1 neuron in the external layer. The input vector of the network is a vector with 1024 entries that is the training and test images vector in the new dataset of area 1. The objective is training this artificial neural network with the images set of area 1 and the extraction of feature vector with 31 entries for each image in the new dataset of area 1. The considered activation functions on an artificial neural network configured for the hidden and output layer are Sigmoid. The Logistic Regression cost function is utilized, and the appendix equations have been utilized for optimization and training the network. Finally, after complete training of the network, each of the training and test images of the dataset of area 1 is passed from the network, and the feature vector of a hidden layer of the network with 31 entries for each image is extracted. The obtained vector replaces the image after conversion to pixel values, and the dataset is updated.

Then this section and the proposed method stages for areas 2, 3, and 4 are repeated, and the new datasets are created for area 4. Now, for each image in the dataset of area 5 with dimensions of 31×31 pixels, there is a vector with dimensions of 31 pixels in each of the other four areas, and we combine the corresponding vectors of each image in area 5 with that image. Because the images of the dataset of area 5 are color images, and they have three color images. So, we copy each vector of that image in other areas in three sequential channels. Finally, the new combined image with 32×32 pixels is created, and the new and final datasets have been created that figure 10 shows an image example of the final result. All of the stages of the proposed algorithm are performed automatically with the written program.

\begin{figure}[h]
\centering\includegraphics[width=0.3\linewidth]{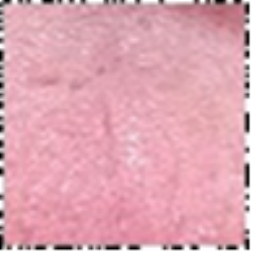}
\caption{A sample image of the final images dataset including a sample image of region 5 and the extracted features of the corresponding images of other regions that are combined  }
\end{figure}

After the generation of the new and final image dataset, the new dataset is trained and tested by deep neural networks.

The main challenge of deep neural networks is the selection of the best architecture and the best network layer for feature extraction from the datasets. Usually, it performs experimentally. In our proposed method, we have selected information theory, the best architecture, and the best layer to extract the features. Information theory and mutual information help the selection of the layer with the best separability among data classes rather than the next and the previous layer. Indeed, the layer with the least mutual information among the images category shows higher separability of the features, and the layer is selected.

In information theory, mutual information among two or more probability distributions determines the news or information distance of the distributions. Since geometric distances are different from news distance, two distributions may be close in terms of geometric distance but far in terms of information distance or vice versa. We can calculate mutual information for two image matrixes or two feature vectors. Mutual information is a relative criterion of similarity between two probability distributions. The more the mutual information between two probability distributions results in the closer and more similar the distributions.

Therefore, we want to find architecture and a layer of convolutional deep neural network that minimizes mean mutual information between two extracted feature vectors from two image categories, because the reduction of mean mutual information between the feature of two image categories means high information distance of the image categories in the architecture and the layer. In other words, the features of the categories in the especial architecture and the layer are farther apart and more separable. So, the classification of the features of that architecture and the layer is performed more precisely. Considering p(x) and q(x) as probability distribution functions for two separate images of two different categories, the relations of mutual information are as follows.

\begin{equation}
I(x;y)= H(x)-H(x|y)
\end{equation}

\begin{equation}
I(x;y)= H(x)+ H(y) -H(x|y)
\end{equation}

\begin{equation}
H(x)= -\sum P(x)\log P(x)~~ \textit{Shanon Entropy}
\end{equation}

\begin{equation}
H(x)= \frac{1}{1-\alpha} \log \sum_x P(x)^\alpha ~~ \textit{Renyi Entropy}
\end{equation}

The relationship between 4 and 6 can be concluded:

\begin{equation}
\begin{split}
I(x;y)=& -\log P(x)-(-\log P(x|y)) \\
=& \log P(x|y) - log P(x)  \\
=& \log \frac{P(x|y)}{P(x)}  \\
\end{split}
\end{equation}

The relationship between 5 and 6 can be concluded:

\begin{equation}
\begin{split}
I(x;y)=& -\log P(x)-\log P(y)- (- \log P(x|y)) \\
=& \log P(x|y) - (\log P(x) - \log P(y)  ) \\
=& \log \frac{P(x|y)}{P(x)P(y)}  \\
\end{split}
\end{equation}

Mutual information between two images or two feature vectors can be calculated by equation 9.

\begin{equation}
P(x|y)=P(x)P(x|y)=P(y)P(y|x) 
\end{equation}

Using Shannon entropy and the following equations, it can be proved that mutual information in figure 11 is an average divergence that is news and information distance.

\begin{equation}
D(P(x)||Q(x))= -\sum_x P(x)\log \frac{P(x)}{Q(x)} 
\end{equation}

Kullback–Leibler divergence of Shannon entropy.

\begin{figure}[h]
\centering\includegraphics[width=0.5\linewidth]{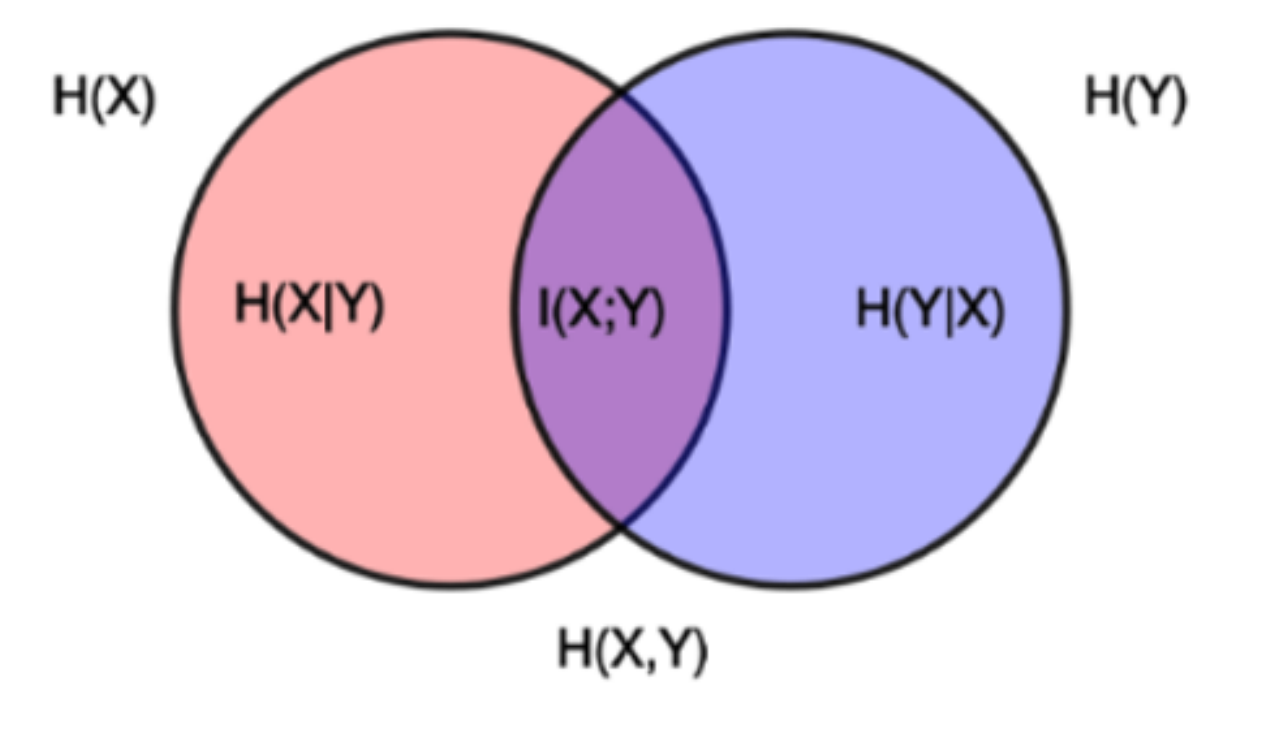}
\caption{Display of mutual information between two possible distributions  }
\end{figure}

Using mutual information between the training images feature vectors in different layers of different architectures of deep neural networks, we can select some layers of the network with more precision and test them. Our prediction about the layers with better precision for our dataset came true after tests. The results have been explained in the following section. The obtained precision of tongue image sets classification in two classes of the patient and healthy using the proposed methods and DenseNet deep neural network with nest layer selection using information theory is 90\%. The general diagram of the proposed method and algorithm has been presented in figure 9.

\section{ Results}
In this section, the \color{black} experimental \color{black} results have been \color{black} discussed comprehensively \color{black} . To this aim, table 1 shows the characteristics of different convolutional deep networks used in the experiments. As it is shown, Alexnet and DenseNet networks have least and most depth, respectively. Moreover, GoogleNet and VGG-16 have least and most number of parameters, include weight and bias of each layer, respectively.

In the following, the classifier precision as a pre-processing of the primary dataset on the pre-trained architectures has been reported in table 2. As it shows, the DenseNet network has the highest precision among other architectures. Table 3 shows classifier precision for different deep architectures on different models for the new dataset and the proposed method. Based on the table, DenseNet has absolute superiority rather than other models. In addition to precision, other criteria like sensitivity, detectability, positive or negative predicted value, and F1-Score have been reported. Based on these criteria, DenseNet results are better than other models. The reported criteria show the percentage of test data that are classified correctly, true-positive diagnosis rate, and true-negative diagnose rate, respectively. Moreover, positive and negative prediction criteria are the related prediction percentage and the percentage of all the predictions that are categorized correctly by the model, respectively. F-Score represents a harmonic average of two previous criteria that is shown in the last column of the tables. Based on the results, all the criteria show the superiority of the DenseNet model rather than other models in recent years.

As it is shown, the F1-Score of DenseNet network is more than other networks that show the superiority and efficiency of the network. Moreover, the positive and negative predictive value of this network has an acceptable superiority rather than other networks that show the true classification of input data. ResNet-152 is the second network with better performance rather than other models. But, its computation time increases because of density and a high number of convolutional layers.

Sensitivity and detectability show the fraction of positive and negative answers that are detected correctly. The value of these statistical indexes that are used for the evaluation of a binary two-state classifier is acceptable.

The important point of the DenseNet model that makes this model different from other models is dense layer connections and simpler and more efficient training rather than other models. The density of layers is because of the dependence of each layer input to the output of all the previous layers.

Table 1. Characteristics of tested convolutional deep neural networks
Parameters (Millions)	Depth	Network

\begin{table}[h!]
\caption{ Characteristics of tested convolutional deep neural  		Network}
\centering
 \begin{tabular}{|c c c |} 
 \hline
 networks & Depth & Parameters (Millions) \\ 
 \hline\hline
  	AlexNet     &8	& 61 \\
    GoogleNet   & 22 & 7\\
 	VGG-16      & 16 & 138\\
    ResNet-50   & 50 & 25.6\\
 	ResNet-101  & 101 & 44.6\\
	ResNet-152  & 152 & 60.3\\
	DenseNet    & 201 &20\\
 \hline
 \end{tabular}
\end{table}


\begin{table}[h!]
\caption{ Classification precision of the primary dataset on different architectures of deep neural networks}
\centering
 \begin{tabular}{|c c c c c c c|} 
 \hline
  Architectures& ACC & SEN & SPE & PPV & NPV & F1-Score \\ 
 \hline\hline
  	AlexNet     & 65\% & 64\% &66\% &69\% &61\% &66\% \\
    GoogleNet   & 65\%& 62\%&71\% &79\% &51\% &70\% \\
 	VGG-16      & 59\%&58\% &62\% &68\% &50\% &63\% \\
    ResNet-50   & 58\%& 56\%&64\% &78\% &39\% &65\% \\
 	ResNet-101  & 64\%&70 \%&61\% &49\% &79\% &58\% \\
	ResNet-152  & 64\%&79\% &59\% &39\% &89\% &52\% \\
	DenseNet    & 69\%& 74\% & 66\% & 59\% & 79\% &66\% \\
 \hline
 \end{tabular}
\end{table}

Note. ACC = accuracy, SEN = sensitivity, SPC = specificity, PPV = positive predictive value, and NPV = negative predictive value.

\begin{table}[h!]
\caption{ Classification precision of new dataset of the proposed method on different architectures of deep neural networks}
\centering
 \begin{tabular}{|c c c c c c c|} 
 \hline
  Architectures& ACC & SEN & SPE & PPV & NPV & F1-Score \\ 
 \hline\hline
  	AlexNet     & 70\% &65 \% &83\% &89\% &51\% &75\% \\
    GoogleNet    & 75\% &72 \% &79\% &81\% &69\% &77\% \\
 	VGG-16       & 69\% & 75\% &66\% &58\% &80\% &66\% \\
    ResNet-50    & 70\% & 82\% &64\% &51\% &89\% &63\% \\
 	ResNet-101   & 74\% & 84\% &69\% &61\% &88\% &70\% \\
	ResNet-152   & 79\% & 87\% &75\% &69\% &89\% &77\% \\
	DenseNet     & 91\% &98 \% &85\% &83\% &98\% &90\% \\
 \hline
 \end{tabular}
\end{table}

\section{ Conclusion and Future Works}
Today, cancer is one of the diseases that affect human life. Also, gastric cancer is one of the common kinds of cancer. By the advancement of technology and utilizing artificial intelligence algorithms can diagnose such diseases and help the patients. In this article, a method has been proposed to increase the diagnosis precision of gastric cancer using surface and color features of tongue based on convolutional deep neural networks and support vector machine.  In the first step of the proposed method, the tongue area is separated from the input images using deep neural networks. Then after required preprocessing, data is transmitted to the convolutional neural network, and training and test are performed. The results show that the proposed method can detect a high precision rate of tongue area and distinguish patients and non-patients. Moreover, based on the examinations, the DenseNet network has the highest precision, among other deep architectures. As the future works can consider the following cases: using evolutionary algorithms to select the optimized number of layers and filters and using Generative Adversarial Networks (GANs) to data production and precision improvement of the model.


\bibliographystyle{elsarticle-num-names}
\bibliography{sample.bib}

\end{document}